\documentclass[11pt,a4paper]{article}
\usepackage[hyperref]{acl2021}
\usepackage{latexsym}
\usepackage{tgtermes}
\usepackage{graphicx}
\usepackage{float}
\usepackage{url}
\usepackage{array}
\usepackage{url}
\usepackage[T1]{fontenc}
\usepackage{lipsum}
\usepackage{amsmath}
\usepackage{amssymb}
\usepackage{booktabs,tabularx}
\usepackage{verbatim}
\usepackage{pgfplots}
\usepackage{multirow}
\usepackage{microtype}
\usepgfplotslibrary{statistics}
\pgfplotsset{compat=1.16}
\newcolumntype{Y}{>{\raggedright\arraybackslash}X}
\newcolumntype{P}[1]{>{\centering\arraybackslash}p{#1}}
\usepackage{caption}
\usepackage{subcaption}
\usepackage{adjustbox}

\aclfinalcopy 

\title{Exploring Story Generation with Multi-task Objectives in 
Variational Autoencoders}

\author{Zhuohan Xie 
 \qquad
 Trevor Cohn 
 \qquad
 Jey Han Lau \\
 School of Computing and Information Systems \\
 The University of Melbourne \\
 zhuohanx@student.unimelb.edu.au, 
t.cohn@unimelb.edu.au, 
 jeyhan.lau@gmail.com }

\date{}

\begin{document}
\maketitle
\begin{abstract}
GPT-2 has been frequently adapted in story generation models as it provides powerful generative capability.
However, it still fails to generate consistent stories and lacks 
diversity.
Current story generation models leverage additional information such as plots or commonsense into GPT-2 to guide the generation process.
These approaches focus on improving generation quality of stories while 
our work look at both quality and diversity. We explore combining BERT 
and GPT-2 to build a variational autoencoder (VAE), and extend it by 
adding additional objectives to learn global features such as story 
topic and discourse relations. 
Our evaluations show our enhanced VAE can provide better quality and 
diversity trade off, generate less repetitive story content and learn a 
more informative latent variable.
\end{abstract}

\section{Introduction}

Autoregressive pretrained models such as GPT-2 
\citep{radford2019language} have been frequently applied to story 
generation.
While GPT-2 can generate coherent single sentences, it suffers from 
inconsistencies in the storylines and lacks \textit{generation 
diversity}, i.e.\ the storylines tend to use ``bland'' language and 
multiple generation produces similar plot lines \citep{guan-etal-2021-long}.
Current story generation models add more controllability into language 
models for story generation, such as story plan \citep{yao2019plan} or 
commonsense \citep{guan2020knowledge}.
These approaches focus on improving generation quality but does not 
address the diversity issue.

Variational autoencoder (VAE) is an extension of autoencoder (AE) 
\citep{10.5555/104279}.
It defines a prior distribution and the encoder learns an approximate 
posterior distribution that is optimised close to the prior 
distribution.
In doing so, the VAE is able to learn a more tractable latent space than 
AE and it is easier to sample meaningful latent variables to guide the 
generation process to generate diverse meaningful sequences.

In order to leverage pretrained models for VAE,
\citet{li-etal-2020-optimus} propose OPTIMUS, a large-scale VAE that 
combines BERT \citep{devlin-etal-2019-bert} and GPT-2 
\citep{radford2019language} and further pretrain it on large corpus to 
create an off-the-shelf pretrained VAE.
We follow a similar approach to build our VAE in this paper, but our aim 
is to develop a VAE for domain-specific story generation (rather than 
creating a domain-general large-scale pretrained VAE) and as such our 
evaluation focuses on assessing generation capability.

Our core innovation in this paper is the introduction of multi-task 
learning objectives to the VAE to enhance the latent variables, as 
\citet{bosc-vincent-2020-sequence} found that they tend to learn local 
features such as the first few words or the length of input sequences.
Our first auxiliary objective uses the latent variable to learn story 
topics, and our second objective seeks to distinguish between original 
stories and ``negative samples'', created by altering the stories to 
simulate common machine generation errors.
We conduct experiments on several datasets to show our proposed VAE has 
better quality-diversity trade off than GPT-2 and learn better latent 
representations than vanilla VAE.

To summarise: (1) we combine BERT and GPT-2 to build domain-specific VAE 
for story generation; (2) we propose an alternative approach to 
incorporate the latent variable into the VAE's decoder; (3) we introduce 
two auxiliary objectives to encourage the latent variable to capture 
topic information and discourse relations; and (4) we experiment with 
several story datasets and show that our enhanced VAE produces higher 
quality latent variables and generates stories with better 
quality-diversity trade off compared to GPT-2.

\section{Related Work}

Conventional approaches of automatic story generation typically contain 
two parts: 
(1) learn a language model from the training dataset with the objective 
of minimising KL divergence between probability distribution of training 
dataset and language model; and (2) find the most suitable way to decode 
the story from a given starting point (usually a title or the leading 
context) with the trained language model.
Autoregressive transformers such as GPT-2 \citep{radford2019language} and its scaled-up GPT-3 \citep{brown2020language} mask the attention heads after the current word during training so that they can serve as language models to predict the next token.
However, even large pretrained language models suffer from issues such 
as self-repetition, conflicting logic and incoherence 
\citep{guan-huang-2020-union}. 

Therefore, recent approaches resort to two main strategies to alleviate 
above issues, by adding more controllability into the story generation 
model and incorporating commonsense knowledge.
One of the most influential strategies of controllability is ``plan and write" \citep{yao2019plan} where they first use a RAKE algorithm to extract the most important word from each sentence and train a storyline planner based on such dataset.
The language model is trained conditional on both the previous context 
and the keywords.
During generation, the keywords are generated from the given title and can be used to guide generation of each sentence.
Commonsense contains shared 
knowledge about the world \citep{alabdulkarim-etal-2021-automatic}.  
\citet{guan2020knowledge} fine-tune a pretrained GPT-2 with knowledge 
triples from commonsense datasets.  They first use pre-defined rules to 
turn triples into sentences (e.g.\ \texttt{(eiffel tower, AtLocation, 
paris)} $\rightarrow$ ``eiffel tower is at paris'') and train on the 
knowledge sentences with conventional maximum likelihood estimation 
objective.  \citet{xu-etal-2020-megatron} combine these two approaches 
by first training a keyword planner with GPT-2 and use the keywords to 
search a knowledgebase to retrieve the top ranked sentences to guide the 
generation process.

The aforementioned approaches add complementary information in training 
the language model, but does not address the diversity issue in language 
generation.
VAE can generate content with more diversity 
\citep{kingma2019introduction, Yu_Li_Liu_Zhao_Yan_Tang_Zhang_2020}, and 
has been variously explored in story generation. For example,
\citet{jhamtani-berg-kirkpatrick-2020-narrative} treat the latent 
variables as story plots to guide story generation and
\citet{Yu_Li_Liu_Zhao_Yan_Tang_Zhang_2020} build a hiererchical 
conditional VAE draft and edit stories.

To incorporate pretrained models for building VAEs,
\citet{li-etal-2020-optimus} propose OPTIMUS, a VAE that uses BERT 
\citep{devlin-etal-2019-bert} as the encoder and GPT-2  
\citep{radford2019language} as the decoder.  They further pretrain 
OPTIMUS on English Wikipedia using standard VAE objectives to create an 
off-the-shelf pretrained VAE, and demonstrate its benefits as a 
pretrained model for downstream tasks. We follow their approach of using 
BERT and GPT-2 for building a VAE, although with a different goal: here 
we are interested in developing domain-specific story generators, and as 
such our evaluation metrics focus on assessing story generation 
capabilities.

Story evaluation is a challenging problem. BLEU 
\citep{papineni-etal-2002-bleu} and ROUGE \citep{lin-2004-rouge} are 
commonly used to assess the quality of generated stories.  Diversity of 
generated stories is another important evaluation aspect and 
\citet{DBLP:conf/iclr/CacciaCFLPC20} propose temperature sweep to 
evaluate the trade off  between quality and diversity for story 
generation models.





\section{Framework}
\label{sec:framework}

Denoting the text sequence as $x$ and the latent variable as $z$, a VAE 
uses the inference model (i.e.\ the stochastic encoder) $q_\phi(z|x)$ to 
approximate the posterior distribution, $p_\theta(z|x)$, since the true 
posterior density $p_\theta(z|x) = p_\theta(x|z)p_\theta(z)/p_\theta(x)$ 
is intractable \citep{Kingma2014}.  The prior over $z$ is set as a 
multivariate Gaussian $p_\theta(z) = N(z;0,I)$.
VAE is trained with the evidence lower bound (ELBO) loss:
\begin{equation}
\label{equ:elbo}
\mathbb{E}_{q_{\phi}(z|x)}[\textup{log}p_\theta(x|z)] - D_{KL} (q_{\phi}(z|x)||p(z))
\end{equation}

The left part of equation can be interpreted as the reconstruction loss 
($L^R$) and the right part as the KL loss ($L^{KL}$) that pushes the 
latent space close to the pre-defined prior so as to obtain a regular 
latent space.

\begin{figure}[t]
    \centering
    \includegraphics[width=0.4\textwidth]{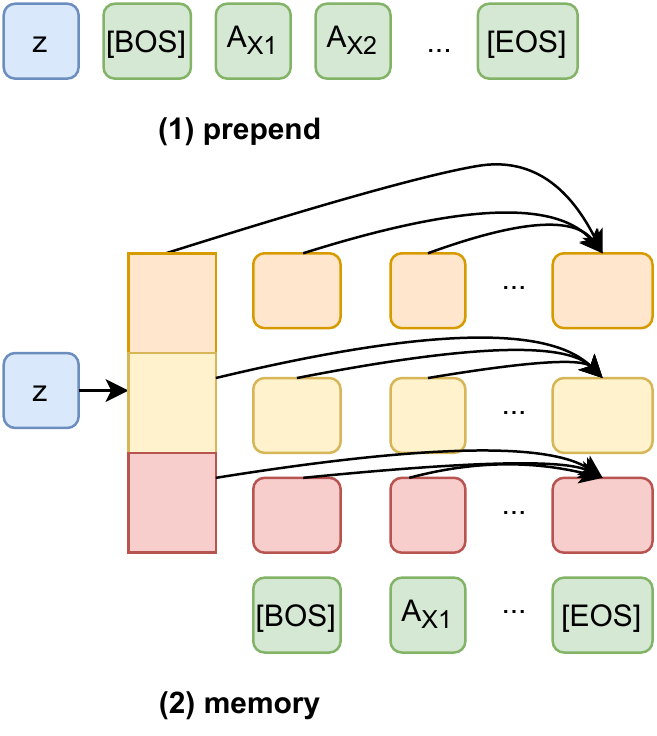}
    \caption{Illustration of three approaches of interacting the latent 
variable with decoder input. Both [BOS] and [EOS] are <|endoftext|> 
token in GPT-2. ``A'' denotes the first sentence of the story, x1 and x2 
represent tokens of the first sentence. For the ``memory'' approach, 
different colors indicate different layers in GPT-2.}
    \label{fig:latentvariable}
\end{figure}

We use BERT as the encoder and GPT-2 as the decoder to build a VAE 
language model.
BERT naturally handles multiple sentences (delimited by [SEP]) and we 
use the [CLS] token to represent the whole story and add two linear 
layers on top to compute the mean ($\mu$) and standard deviation 
($\sigma$) of the latent variable $z$.
To incorporate the latent variable $z$ into the GPT-2 decoder, we 
explore two approaches:
(1) ``prepend'', where we append the latent variable as prefix token at 
the beginning of input sequence. 
(2) ``memory'', where we apply an MLP to the 
latent variable to generate key and values in each layer (proposed by 
\citet{li-etal-2020-optimus}); and
Figure \ref{fig:latentvariable} presents an illustration of these two 
approaches.


\subsection{Global Feature Learning}
\label{sec:global-feature-learning}

\begin{figure}[t]
    \centering
    \includegraphics[width=0.5\textwidth]{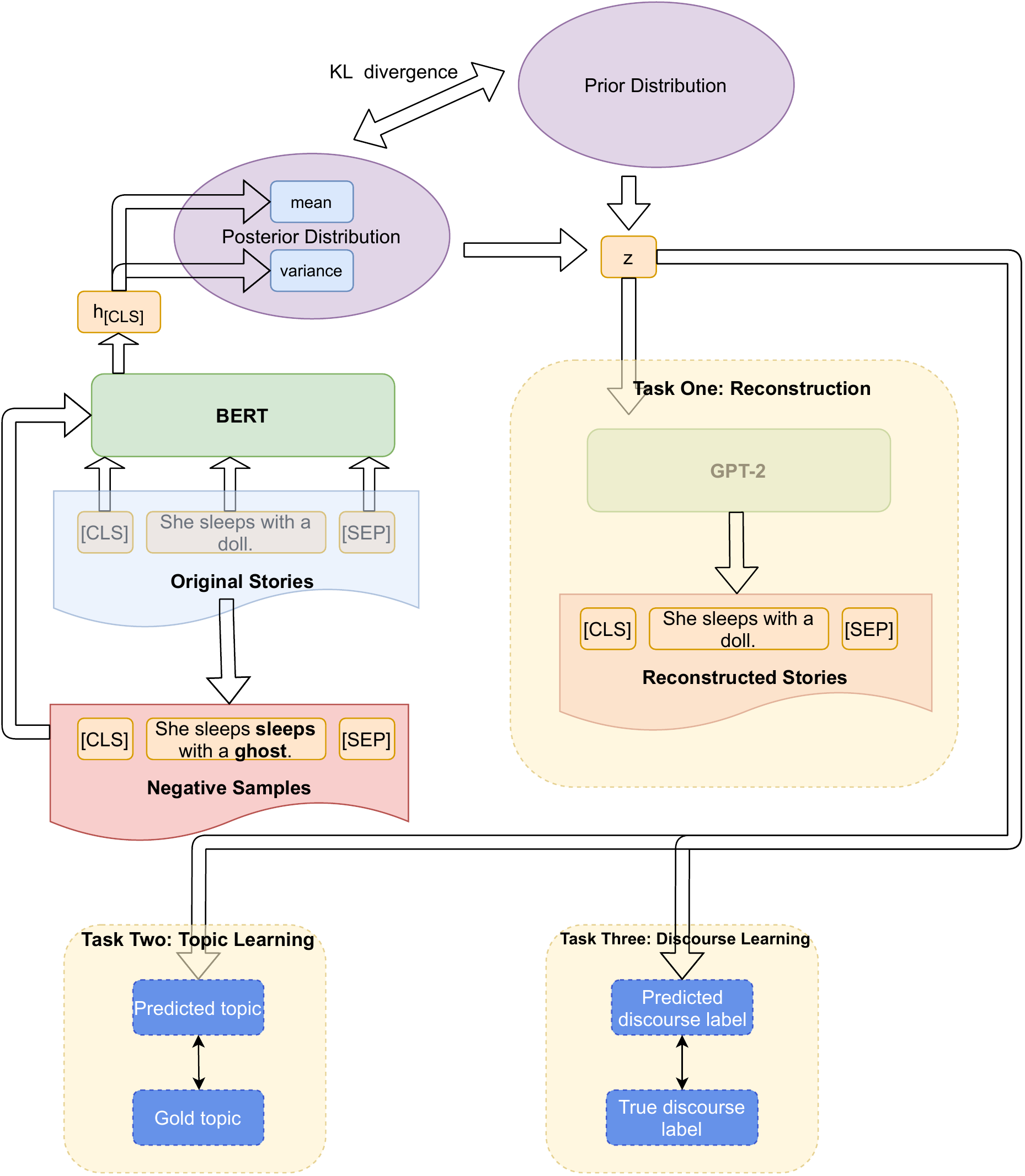}
    \caption{Our proposed multi-task VAE model with pretrained BERT and 
    GPT-2. In additional to the original objective of reconstructing the 
original story, the latent variable is also used to predict the story 
topic and distinguish between original and negative samples.  Here we 
show a simple one sentence story and the negative sample is constructed 
using repetition and substitution. Our training dataset contains stories 
of multiple sentences, separated by [SEP].}
    \label{fig:multi-task}
\end{figure}

To encourage the VAE to learn global features, we propose a multi-task 
learning framework. Figure \ref{fig:multi-task} presents an overall 
architecture of our model.
The first objective is the reconstruction objective ($L^R$, the left 
part of Equation \ref{equ:elbo}). The two additional objectives train 
latent variable to: (1) predict the story topic; and (2) distinguish 
between negative samples vs. original stories.
These auxiliary objectives are designed to encourage the latent variable 
to capture topic and discourse information.

\paragraph{Story Topic Learning}
We add additional MLP layers to learn the topic distribution of the 
story and calculate the topic loss with the ground truth topic 
distribution of the document based on KL divergence.  While this is  
straightforward for topic-annotated dataset which contains ground truth 
topic labels, most story datasets do not have such label.
To this end, we train a latent Dirichlet allocation topic model 
\citep{LDA} to extract the topics.
We use the topic model-inferred topic distribution $Q(T)$ of each document as ground truth and compute
KL divergence as the loss. Note that we use the full topic distribution 
instead of selecting one topic with the highest probability as the 
representative topic as the full distribution is more informative and 
that most documents have multiple topics.

Given $z$, we predict the topic distribution $P(T)$ as follows:
\begin{equation}
\label{equation:topicloss1}
P(T) = \textup{softmax}(W_tz + b_t)
\end{equation}

We calculate the topic loss $L^T$ with KL divergence over the predicted and topic model-inferred topic distribution as follows:

\begin{equation}
\label{equation:topicloss2}
L^T = \sum_{t \in T }P(t)\textup{log}\left ( \frac{P(t)}{Q(t)}\right )
\end{equation}


\paragraph{Story Discourse Learning}

For discourse relation learning, we first construct negative samples from the original stories.
Following \citet{guan-huang-2020-union}, we use a random combination of 
four heuristic rules to construct the common machine generation issues 
in story: (1) repeat of n-grams or sentences, (2) substitution of random 
keywords or the entire sentence, (3) reordering of sentences and (4) 
negation alteration of the original sentences. Table 
\ref{table:negativeexamples} presents some examples of original stories
and altered stories (negative samples).

\begin{table*}[t]
\centering
\begin{tabular}{p{2cm} p{6cm} p{6cm}} 
 \toprule
 \textbf{Rule} & \textbf{Original Story} & \textbf{Negative Sample}  \\ [0.5ex] 
 \midrule
 repeat, substitution and negation alteration  & [NEUTRAL] knew the solution to a problem . he told people the solution . the people thought [NEUTRAL] was smart . [NEUTRAL] agreed with them . [NEUTRAL] went on to achieve . & [NEUTRAL] knew the solution to a problem . he told \textbf{animals} the solution . \textit{[NEUTRAL] did not go on to achieve . he told animals the solution . he told animals the solution .} \\
 \midrule
 reordering and substitution & [FEMALE] really loved the sun . she would play in it all day . one day the dark clouds came and shooed the sun away . [FEMALE] was very sad to see it go . she was happy though when she saw it back the next morning ! & 
 [FEMALE] really loved the sun . \textit{she was happy though when she saw it back the next morning !} she would play in it all day . [FEMALE] was very sad to see it go . one day the dark clouds came and shooed the \textbf{moon} away . \\
 \bottomrule
\end{tabular}
\caption{Examples of negative story samples generated from a combination of heuristic rules of repeat, substitution, reordering and negation alteration.}
\label{table:negativeexamples}
\end{table*}

Given a story and its discourse label (1.0 for original stories or 0.0 
for  negative samples) and $z$,
we apply a linear layer on $z$ to compute the discourse score 
$\hat{y}_n$:
\begin{equation}
\label{equation:disloss1}
\hat{y}_n = \textup{sigmoid}(W_dz + b_d)
\end{equation}

We then compute the discourse loss $L^D$ using standard binary cross 
entropy:
\begin{equation}
\label{equation:disloss2}
L^D = -y_n\textup{log}\hat{y}_n - (1-y_n) \textup{log}(1-\hat{y}_n)
\end{equation}

For each original story, we create one negative sample.
%

Given the topic and discourse losses, we add them with weights to the 
original reconstruction loss and KL loss function to train the VAE and 
perform grid search to find the suitable weights.
During training, to alleviate posterior collapse --- the issue where 
both the variational posterior distribution obtained from the encoder 
and the true posterior for the real dataset collapse to the prior, 
resulting in zero KL loss \citep{lagging} --- we use $\beta$-VAE 
\citep{burgess2018understanding}
that sets an additional target $C$ to the KL loss (by computing an  
absolute difference between KL loss and $C$) to optimise it close to 
$C$.  The full objective our model is thus given as follows:
\begin{equation}
\label{equation:allloss}
L = L^R + \beta\ \left | L^{KL} - C  \right | + \alpha\ L^T + \gamma\ L^D
\end{equation}
where $\beta$, $\alpha$ and $\gamma$ are hyper-parameters to control the 
weights of different objetives.


\section{Dataset}

We use four datasets in our experiments: ROCStories, APNEWS, Reuters and 
WritingPrompts. APNEWS is a collection of Associated Press news \citep{bhatia-etal-2017-automatic}  from 2009 to 2016.
Reuters\footnote{https://www.kaggle.com/nltkdata/reuters} is the 
Reuters-21578 ``ApteMod" corpus for text categorization from the Reuters 
financial newswire service.  ROCStories (ROC) contains commonsense 
stories of five sentences \citep{mostafazadeh-etal-2016-story}. To 
obtain more generalization as all sentences are rather short in the dataset, we follow the 
delexicalization approach from prior studies \citep{guan2020knowledge, 
xu-etal-2020-megatron} where male/female/unknown names are replaced by 
tokens [MALE]/[FEMALE]/[NEUTRAL].
The WritingPrompts (WP) dataset consists of 303,358 human generated long 
stories from Reddit's Writing Prompts
forum\footnote{https://www.reddit.com/r/WritingPrompts/}.
\citet{fan-etal-2018-hierarchical} collect them by scraping three years 
of prompts and their associated stories.
We use 10\% of the stories in our experiments.
Table \ref{table:dataset} presents some statistics of the four datasets.  

In terms of preprocessing, we add [SEP] token at the end of each 
sentence and use WordPiece tokenizer  for BERT and Byte-Pair-Encoding 
(BPE) for GPT-2.
We set the maximum length of a story as $100$ subwords for short story 
datasets (ROC and Reuters) and $200$ for long story datasets (APNEWS and 
WritingPrompts).

\begin{table*}[t]
\centering
\begin{tabular}{c c c c c c c c } 
 \toprule
 \multirow{2}{4em}{\textbf{Collection}} & \multirow{2}{4em}{\textbf{Average Length}} & \multicolumn{2}{c}{\textbf{Training}} & \multicolumn{2}{c}{\textbf{Development}} & \multicolumn{2}{c}{\textbf{Test}}  \\
 & & \#Docs &\#Tokens &\#Docs & \#Tokens & \#Docs& \#Tokens   \\
 \midrule
 APNEWS & 138 & 46.4K & 4.68M & 1.9K & 187K & 1.8K & 187K  \\ 
 Reuters & 88 & 7.8K & 695K & 2K & 180K & 1K & 93.6K  \\
 ROC & 60 & 88K & 5.28M & 5K & 0.3M & 2K & 0.12M  \\
 WritingPrompts & 110 & 26.8K & 2.95M & 2K & 0.22M & 2K & 0.22M  \\
 \bottomrule
\end{tabular}
\caption{Statistics of APNEWS, Reuters, ROC and WritingPrompts Dataset. }
\label{table:dataset}
\end{table*}

\section{Experiments}
\label{subsec:experiments}

We use implementations of BERT and GPT-2 from HuggingFace \citep{huggingface}. 
We set learning rate at $10^{-4}$ and use Adam \citep{adam} as optimiser. 
The dimension of latent variable is set as $256$.
All models are trained using $20$ epochs on single NVIDIA V100 GPU node per model.
\subsection{Topic Extraction}
We use MALLET LDA\footnote{http://mallet.cs.umass.edu} to extract the 
topics.
We filter out tokens that appear more than half of the dataset and keep 
the most frequent 50K tokens as the vocabulary for the LDA models.
We select the best topic number based on topic coherence 
\citep{Rder2015ExploringTS}.


\subsection{Evaluation Metrics}

We evaluate our system using intrinsic metrics where we compute 
perplexity, number of active units of language model training and the 
extent to which the latent variable captures topic and discourse 
information.
To evaluate story generation capability, we look at self-repetition 
metrics and measure the quality-diversity trade off using Corpus-BLEU.

\paragraph{Perplexity (PPL)}
Perplexity of test data is widely used to evaluate language models.  
However, exact PPL is unavailable so ELBO is often used to approximate 
the probability.  But as \newcite{li-etal-2019-surprisingly} found, such 
approximation is not appropriate since the gap between ELBO and log 
marginal likelihood might be large when the true posterior did not 
converge with the approximate posterior.  \newcite{activeunits} propose 
using $k$-sample importance weighting estimate, which provides a tighter 
lower bound for the log marginal likelihood with Jensen's inequality.  
Our results therefore use this approach for computing PPL. 

\paragraph{Number of Active Units (AU)}
\citet{activeunits} propose a way to evaluate if each dimension of the latent variable is active over the posterior distribution as follows:
\begin{equation}
\label{equation:au}
A_{u} = \textup{Cov}_{x}(\mathbb{E}_{u \sim q(u|x)}[u])
\end{equation}
and set the bar that the dimension $u$ of the latent variable is active 
if $A_{u} > 0.01$. Intuitively, more active units means a more 
informative latent variable is learned from the input.

\paragraph{Sequence Repetition}
As neural generation models are prone to generate repetitive content with high probabilities \citep{DBLP:journals/corr/abs-1811-05701},
we evaluate sequence-level repetition evaluation by computing the 
portion of duplicate n-grams for a continuation $x_{k+1:k+N}$: \begin{equation}
\label{equation:rep-seq-n}
 1.0 - \frac{|\textup{unique n-grams}(\mathbf{x}_{k+1:k+N})|}{|\textup{n-grams}|}
\end{equation}

\paragraph{Corpus-BLEU and Self-BLEU}
Corpus-BLEU uses the test dataset as reference and compute BLEU score 
for each
generated story and use average result as a measurement of quality. 
\citet{zhu2018texygen} propose Self-BLEU, that regards one generated 
story as the hypothesis and all other generated stories as the 
references and calculates the BLEU score for each story and use the 
average score to measure diversity. A lower Self-BLEU score means the 
story is less similar to the other generated stories, and thus, higher 
diversity.

\subsection{Evaluation Results}
\subsubsection{Intrinsic Results}
\label{subsec:results}
We first show evaluation results where we explore two methods 
(``memory'' and ``prepend'') of injecting $z$ to the decoder on ROC in 
Table \ref{table:testresult}. Here the models are vanilla VAE models 
without the auxiliary losses (as our objective here is to evaluate the 
best way to incorporate the latent variable to the VAE's decoder). Note 
that perplexity is estimated using $500$ samples with importance 
weighting and it captures both reconstruction and KL loss.
We found that ``prepend'' generally outperforms ``memory'', as it can 
keep more dimensions of the latent variable active while ``memory'' has 
no active dimensions.  It also has a KL divergence marginally closer to 
the target ($C$ in Equation \ref{equation:allloss}), and has better 
reconstruction and overall better perplexity.
``prepend'' is in a way similar to memory where all tokens in the GPT-2 
input have the extra vector to attend to, but instead of transforming it 
using extra MLP layers, ``prepend'' relies on the inherent 
self-attention mechanism to produce a more natural key/value 
representations in each layer, which might explain the improved 
performance.


By increasing $C$ for the KL target, more information is encoded into 
the latent variable, and so the model achieves a better performance in 
terms of reconstruction loss.
But this also means it becomes harder to sample a latent variable from 
the prior, as the posterior no longer matches the prior, and as such we 
see an increase of perplexity. Our results highlight the importance of 
controlling $C$ to find a reasonable trade off between reconstruction 
and KL loss.

\begin{table}[t]
\centering
\begin{adjustbox}{max width=\linewidth}
\begin{tabular}{crcccc}
\toprule
\textbf{Method} & $\mathbf{C}$ &  \textbf{Recon. loss} & \textbf{KL 
loss} & \textbf{AU} & \textbf{PPL} \\
\midrule
 prepend & 6.0 & 123.89 & 5.96 & 209 & 9.53 \\ 
 prepend & 8.0 & 122.61 & 7.99 & 206 & 9.58 \\
 prepend & 10.0 & 121.64 & 9.96 & 197 & 9.61 \\ 
 memory & 6.0 & 127.67 & 5.94 & 0 & 9.69 \\ 
 memory & 8.0 & 127.49 & 7.93 & 0 & 9.80 \\
 memory & 10.0 & 127.46 & 9.84 & 0 & 9.96 \\
\bottomrule
\end{tabular}
\end{adjustbox}
\caption{Intrinsic results of training with different $C$ in beta-VAE 
(Equation \ref{equation:allloss}) and with ``prepend'' and ``memory'' 
(Section \ref{sec:framework}) for incorporating the latent variable to 
the decoder on the ROC dataset.  PPL is computed by $500$ samples of 
importance weighting estimate. }
\label{table:testresult}
\end{table}

Given these results, we next train the VAE with the topic and discourse 
objectives (Section \ref{sec:global-feature-learning}), using $C = 6.0$ 
and the ``prepend'' method. 
We now assess the extent to which the encoder can identify the topics or 
distinguish between the original stories and stories with flaws 
(negative samples).

\paragraph{Topic Learning Evaluation}

We evaluate the extent to which the BERT encoder can learn story topics 
in the latent space and how much the GPT-2 decoder can make use of it.  
We use the Reuters dataset here since the documents/stories are 
annotated with ground truth topics. 

We follow \citet{bosc-vincent-2020-sequence} and freeze the parameters 
of BERT and add one MLP layer on top of the mean of the posterior distribution 
$\mu$ and the latent variable $z$ and train a classifier to predict the 
ground truth topics 
and report test accuracy results in Table \ref{table:classifier1}. The 
baseline ``AE'' is a VAE model without using the KL loss ($L^{KL}$ in 
Equation \ref{equ:elbo}), and so functions like an autoencoder (since 
the posterior is no longer constrained to be close to the prior).

Looking at the results, we see that using $\mu$ as input for the 
classifier yields much better results compared to using the latent 
variable $z$. But as pointed out in \citet{bosc-vincent-2020-sequence}, 
$z$ is ultimately the latent variable that goes into the decoder, and so 
the performance using $z$ is the more important number.  There is no 
surprise that AE achieves better test accuracy scores with both $\mu$ 
and $z$ than 
vanilla VAE since the VAE's encoder is forced to discard some 
information in the posterior distribution so as to match the prior 
distribution.
Encouragingly, we see that our topic-enhanced VAE is indeed able to 
capture much of the topic information, producing a better topic 
classification accuracy compared to vanilla VAE.

\begin{table}[t]
\centering
\begin{tabular}{c c c } 
 \toprule
 \textbf{Model} &  $\mathbf{\mu}$ & $\mathbf{z}$  \\ [0.5ex] 
 \midrule
 AE & 0.702  & 0.699 \\ 
 VAE & 0.446 & 0.436  \\
 VAE+t & 0.691 & 0.583  \\
 \bottomrule
\end{tabular}
\caption{Topic classification accuracy using mean of the posterior 
distribution $\mu$ and the latent variable $z$ on Reuters.  }
\label{table:classifier1}
\end{table}

\begin{table*}[t]
\centering
\small
\begin{tabular}{c  p{12cm}p{2cm}} 
 \toprule
 \textbf{Score} & \textbf{Story} & \textbf{Issue} \\ [0.5ex] 
 \midrule
 0.83 & [MALE] went fishing . he was excited about the trip . he saw a big fish . he was excited to get it . he caught a huge fish . & \\
 \midrule
0.81 & [FEMALE] was nervous for her first day of school . she was nervous because she was so new to school . [FEMALE] was scared to be in the classroom . the teacher introduced her to other students . [FEMALE] was very excited to learn about her new class . &\\
\midrule
0.56 & [FEMALE] was hungry for some cookies . she decided to make some chocolate chip cookies . she mixed the ingredients together . \textbf{then she mixed them together .} [FEMALE] was happy to have some cookies . & repeat \\
\midrule
0.48 & \textbf{[FEMALE] was a lesbian . she was in love with [MALE] .} [MALE] was jealous of her . [FEMALE] 's boyfriend cheated on her . [FEMALE] was dumped . & conflict logic \\
\midrule
0.40 & [MALE] received a call from his boss . he had a promotion . \textbf{he took it . he took it anyway . he got it .} & repeat and incoherent \\
\midrule
0.32 & [MALE] grew up on a farm . [MALE] wanted to grow vegetables . \textbf{he was tired of them . [MALE] bought carrots .} he then grew vegetables . & incoherent \\
 \bottomrule
\end{tabular}
\caption{Predicted discourse scores using the discourse-enhanced VAE.}
\label{table:scoreandstory}
\end{table*}

\paragraph{Discourse Learning Evaluation}

One advantage of our discourse-enhanced VAE is that after training  we 
can obtain a discourse score using the output of the additional layer 
(Equation \ref{equation:disloss1}), which tells us the quality of a 
story.
Table \ref{table:scoreandstory} presents the predicted discourse scores 
on a set of generated stories. Note that all stories are generated from 
randomly sampled latent variables.
Looking at the generated stories, we found that stories with high 
discourse scores are generally coherent,
while stories with low scores often have logical or repetition problems.
To quantify this, we compute the average discourse score on test stories 
and their negative samples, and the average scores are $0.75$ and $0.25$ 
respectively,
showing that our discourse-enhanced VAE is able to distinguish between 
original stories and negative samples.

\subsubsection{Extrinsic Results}
 

\paragraph{Quality and Diversity Trade-off}
Quality and diversity of generated stories from a model can be affected 
by decoding strategies.  Therefore, it is difficult to determine which 
model is superior based on a single performance since models that 
achieve high quality score tend to lack diversity 
\citep{DBLP:conf/iclr/CacciaCFLPC20}.
Temperature sweep uses a set of quality and diversity results generated by altering values of temperature 
in temperature sampling, 
and the best model is one that produces the 
best trade off between these two aspects 
\citep{DBLP:conf/iclr/CacciaCFLPC20,Hashimoto+:2019, alihosseini-etal-2019-jointly}.
We follow this evaluation approach and use top-$p$ sampling with varying 
$p$ values as \citet{textdegen} demonstrate that top-$p$ sampling has a 
better control over sampling and produce sequences that have a more 
similar nature with human text than temperature sampling.

\begin{figure*}[t]
     \centering
     \begin{subfigure}[b]{0.44\textwidth}
         \centering
         \includegraphics[width=\textwidth]{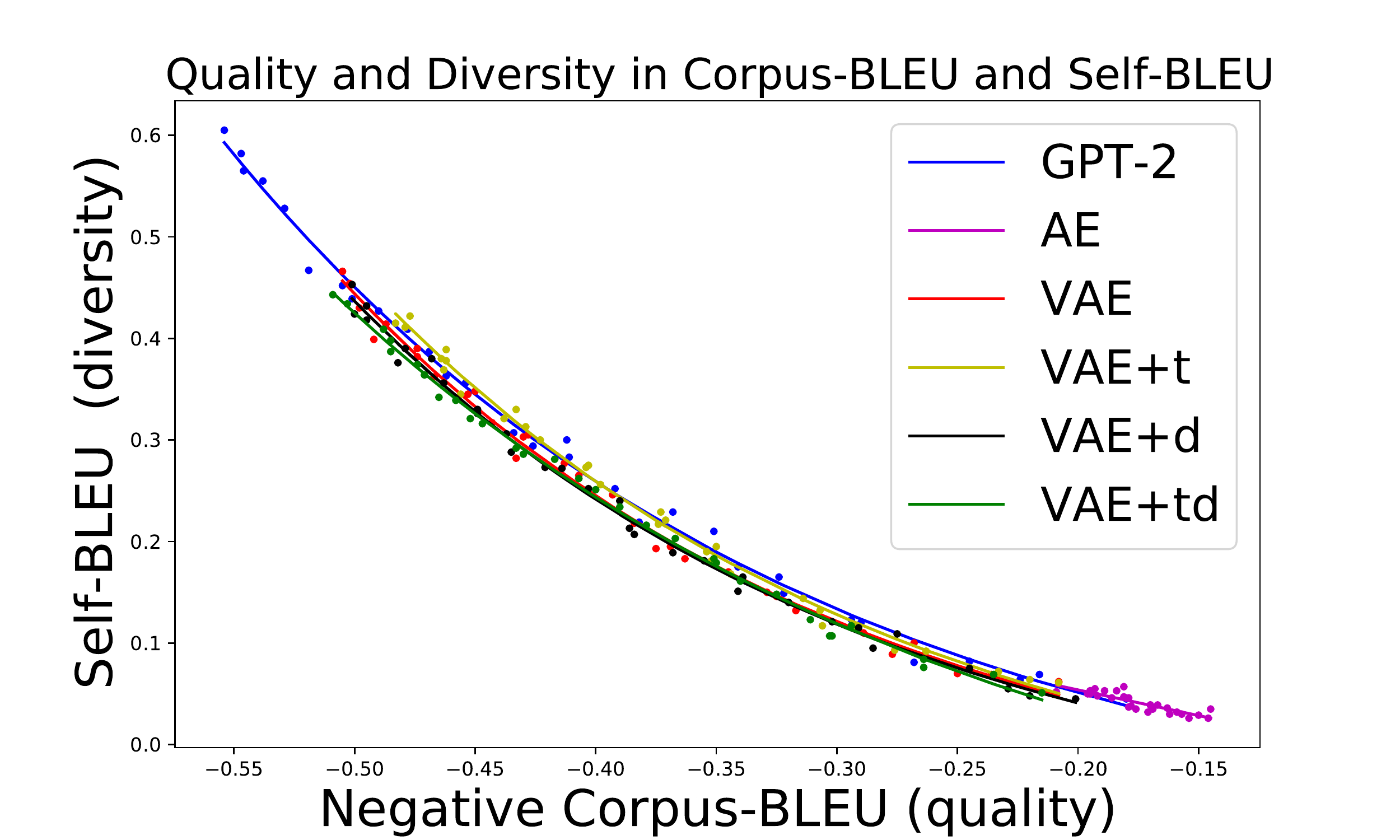}
         \caption{ROC}
         \label{fig:roc}
     \end{subfigure}
     \hfill
     \begin{subfigure}[b]{0.44\textwidth}
         \centering
         \includegraphics[width=\textwidth]{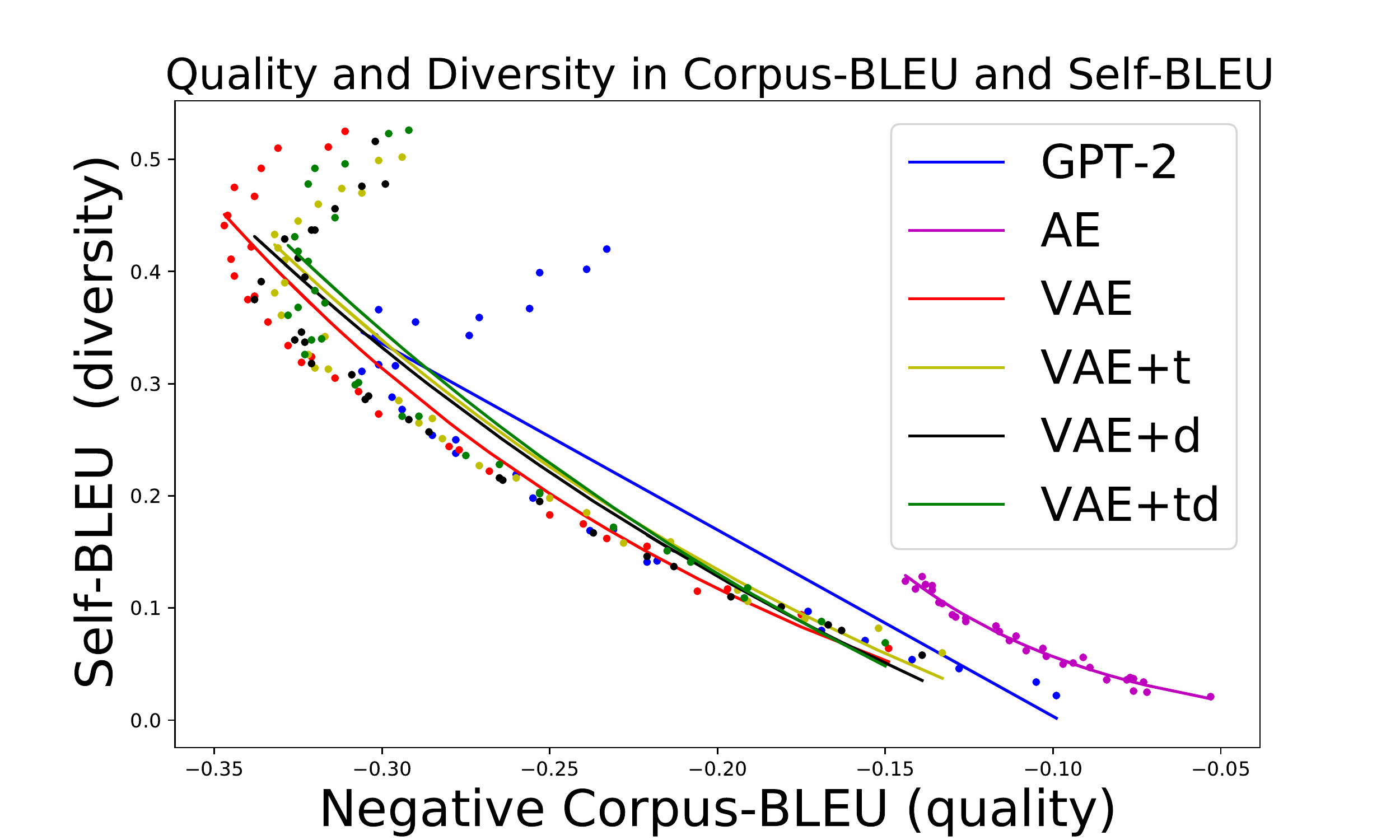}
         \caption{WP}
         \label{fig:wp}
     \end{subfigure}
     \hfill
     \begin{subfigure}[b]{0.44\textwidth}
         \centering
         \includegraphics[width=\textwidth]{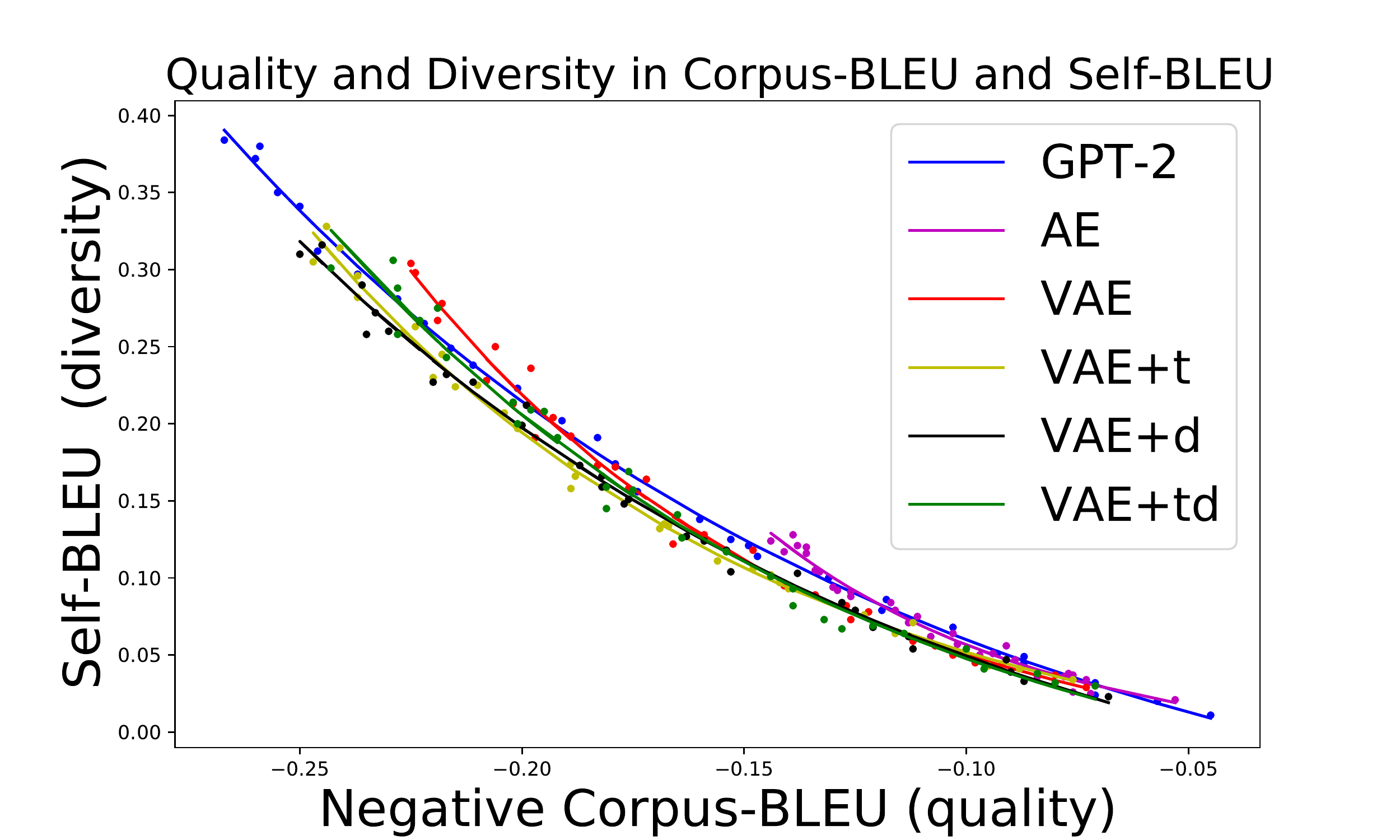}
         \caption{APNEWS}
         \label{fig:apnews}
     \end{subfigure}
        \caption{Quality and diversity trade-offs of generated sentences 
        on three dataset. For both quality and diversity metrics, lower 
score means better performance and the curve that is closest to the axes 
have the best overall performance.}
        \label{fig:generation_qnd}
\end{figure*}

We use a range of different $p$ values from $0.4$ to $1.0$ with an 
increment of $0.02$, creating stories for $31$ different $p$ values to 
assess the quality and diversity trade off.
For each $p$ value, we sample $500$ latent variables from the prior
distribution to generate $500$ stories.
The results are shown in Figure \ref{fig:generation_qnd}. Note that we 
use \textit{negative} Corpus-BLEU here (by flipping the sign), so that a 
lower score indicates better performance for both scores. The best model 
is one that produces a trade off curve closest to the axes.
The figure shows that the VAEs generally achieve a better trade off than 
fine-tuned GPT-2 in all domains.  Encouragingly, our enhanced VAEs 
(``VAE$+$t'', ``VAE$+$d'' and ``VAE$+$td'') also perform generally 
better than the vanilla VAE (with the exception of the WP dataset).
Curiously, AE is not able to generate high quality stories under our 
tested $p$ values and it produces a short curve near the bottom right 
corner.

\begin{table*}[t]
\centering
\begin{tabular}{c c c c c c c c } \toprule
 \multirow{2}{3em}{\textbf{Model}} & \multicolumn{7}{c}{$\mathbf{p}$ \textbf{value}} \\
 & 0.4 & 0.5 & 0.6 & 0.7 & 0.8 & 0.9 & 1.0 \\ 
 \midrule
 GPT-2 & 0.0594 & 0.0300 & 0.0196 & 0.0125 & 0.0081 & 0.0043 & 0.0021 \\ 
 AE & 0.0009 & 0.0006 & 0.0009 & 0.0004 & 0.0004 & 0.0005 & 0.0002 \\
 VAE & 0.0297 & 0.0191 & 0.0153 & 0.0109 & 0.0070 & 0.0042 & 0.0021 \\
 VAE$+$t & 0.0272 & 0.0235 & 0.0185 & 0.0124 & 0.0077 & 0.0045 & 0.0028 \\
 VAE$+$d & 0.0257 & 0.0173 & 0.0143 & 0.0114 & 0.0086 & 0.0049 & 0.0031  \\
 VAE$+$td & 0.0237 & 0.0218 & 0.0168 & 0.014 & 0.0078 & 0.0054 & 0.0031  \\
 \bottomrule
\end{tabular}
\caption{Sequence repetition of $4$-grams of generated stories under 
different $p$ values with top-$p$ sampling on ROC.}
\label{table:seq-rep-4}
\end{table*}


\paragraph{Sequence Repetition}
Self-BLEU measures the diversity of a set of generated stories, 
revealing whether they tend to use similar plots or share similar words.
Here we assess the extent of self repetition \textit{within a story}.  
We compute 4-grams repetition (``seq-rep-4''; Equation 
\ref{equation:rep-seq-n}) and present the
results in Table \ref{table:seq-rep-4} for the ROC 
dataset.\footnote{Other domains produce similar trends and for brevity 
we present only the ROC results.} Note that a lower score means less 
repetition (better performance).

We can see that higher p values produce less repetitive texts (lower 
scores) since at each timestep more word types are included in the 
sampling process.  For comparison, we also compute the ``human'' 
repetition score using the test data and its result is $0.021$.
At lower $p$ values, the VAE models tend to have much lower repetition 
than the fine-tuned GPT-2.
However, if we do not constrain much on the token probabilities and use 
a higher $p$ values, most models produce similar repetition scores.
At the extreme when we set $p = 1.0$, all models are able to generate 
stories with little self-repetition like the human-written stories.
AE seems to be able to repeat less, however the generated stories tend 
to be incoherent (recall in Figure \ref{fig:generation_qnd} we saw it 
has poor Corpus-BLEU scores generally).


\section{Conclusion}

We explore using pretrained models such as BERT and GPT-2 to build a VAE 
for story generation. We additionally propose enhancing the VAE by 
introducing two auxiliary objectives to encourage it to learn topical 
and discourse information in the stories. Our experiments show that the 
latent variable of our enhanced VAE is more informative, in that it 
captures the story topics and  good vs.\ poor quality stories.  In terms 
of story generation, we also demonstrate that our enhanced VAE produce 
generally a better quality-diversity trade off compared to vanilla VAE 
and GPT-2.

\bibliographystyle{acl_natbib}
\bibliography{acl2021}


\end{document}